\documentclass[runningheads]{llncs}

\usepackage{eccv}

\usepackage{eccvabbrv}

\usepackage{graphicx}
\usepackage{booktabs}
\usepackage{multirow}
\usepackage[table]{xcolor}
\usepackage{placeins} %added
\usepackage{subcaption} %added
\usepackage{bm} %added
\newcommand{\ignore}[1]{} %added
\usepackage[accsupp]{axessibility}  % Improves PDF readability for those with disabilities.

\usepackage{hyperref}

\usepackage{orcidlink}

\begin{document}

% ---------------------------------------------------------------
\title{A Classifier-Agnostic Zero-Shot Adversarial Attack Detection via
CLIP} 


% \author{Hodaya Krakover\inst{1}\orcidlink{0000-1111-2222-3333} \and
\author{Hodaya Krakover\inst{1} \and
Meir Yossef Levi \inst{1} \and
Eyal Gofer \inst{1} \and
Guy Gilboa \inst{1}\thanks{This work was jointly supervised by the last two authors.}}

\authorrunning{H. Krakover et al.}

\institute{Technion - Israel Institute of Technology, Haifa, Israel
\email{hodaya.r@campus.technion.ac.il}\\
% \url{http://www.springer.com/gp/computer-science/lncs}
}

\maketitle

\begin{abstract}
  Adversarial attacks pose a challenge to the reliability of deep learning models, motivating effective detection methods. Existing techniques  often rely on attack-specific assumptions, access to adversarial samples, or knowledge of the underlying classifier (white-box). We propose \textit{$A^4D$ (\textbf{A}ttack- and \textbf{A}rchitecture-\textbf{A}gnostic \textbf{A}dversarial \textbf{D}etector)}, a completely black-box, zero-shot adversarial attack detection framework that utilizes prompt-based similarity scores derived from CLIP. To the best of our knowledge this is the first attempt to utilize CLIP for such a task.
  The method is based on two key observations: (i) CLIP is sensitive even to small imperceptible non-semantic perturbations; (ii) The shift in CLIP embedding space is not arbitrary and can be used as a robust attack indicator.
  Experiments across multiple attacks, datasets and classifiers validate that $A^4D$ achieves SOTA detection results in the attack-agnostic and classifier-agnostic setting.
\end{abstract}

\section{Introduction}
\label{sec:intro}

One of the major concerns regarding the reliability of vision encoders is that they can be fooled, sometimes easily, by adversarial attacks \cite{goodfellow2014explaining,madry2017towards,kurakin2018adversarial, croce2020reliable}. While the human eye is largely agnostic to small changes in a scene, even the most advanced vision encoders tend to alter their predictions when small perturbations occur \cite{moosavi2016deepfool}.
With the rise of adversarial attacks, a complementary line of research has emerged, focusing on defenses that aim to enhance the robustness of encoders to such targeted perturbations \cite{lee2018simple, cohen2019certified, shafahi2019adversarial, liang2022adversarial, li2025privacy}. However, given the rapid progress in visual encoders, it is crucial to develop defenses that are generalizable and applicable to a wide range of models, including future ones. Moreover, since there is a wide variety of adversarial attacks, it is desirable for a defense method to remain robust regardless of the specific attack mechanism. In this paper, we focus on a zero-shot \textit{detection} method: given only an input image (and a mini-validation set from the dataset), we aim to determine whether it has been adversarially manipulated or not.

Existing adversarial detection methods typically require additional information. Some approaches rely on attack-specific characteristics and therefore do not generalize well across different attack types \cite{metzen2017detecting, feinman2017detecting, xu2017feature}. Others depend on classifier-specific representations\cite{ma2018characterizing, papernot2018deep, zhang2019defending, danesh2025understanding, stenhuis2025meetsafe}. However, such information may not be available in real-world scenarios. Ideally, we would like to detect attacks without access to the underlying model or the attack mechanism, especially as new encoders and adversarial attacks continuously emerge. In parallel, a large body of work has focused on leveraging foundation models such as CLIP \cite{radford2021learning} for semantic tasks, including detection\cite{zhou2023anomalyclip}, segmentation \cite{peng2025understanding}, classification \cite{wei2023iclip}, and text-to-image generation \cite{crowson2022vqgan}. These models are typically viewed as operating in a semantic latent space, designed for image–text alignment. However, this perspective is incomplete: beyond semantics, CLIP embeddings can also be exploited for non-semantic, low-level vision tasks. In particular, we show that this characteristic can be used for adversarial detection.

\begin{figure}[tb]
    \centering
    
    % ---- First Row ----
    \begin{subfigure}[t]{0.42\linewidth}
        \centering
        \includegraphics[width=\linewidth]{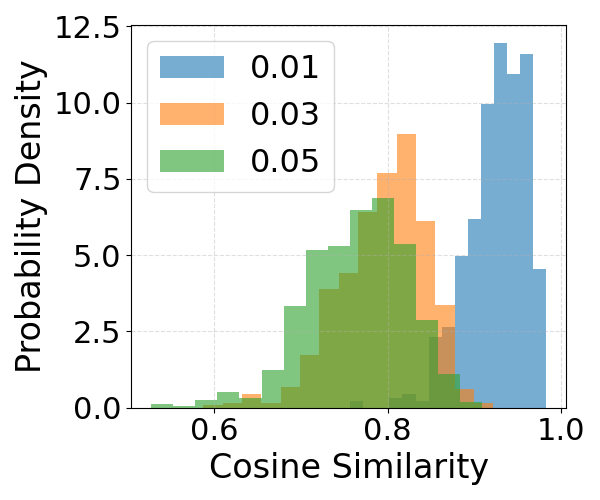}
        \caption{}
        \label{fig:cos_sim}
    \end{subfigure}
    \hfill
    \begin{subfigure}[t]{0.42\linewidth}
        \centering
        \includegraphics[width=\linewidth]{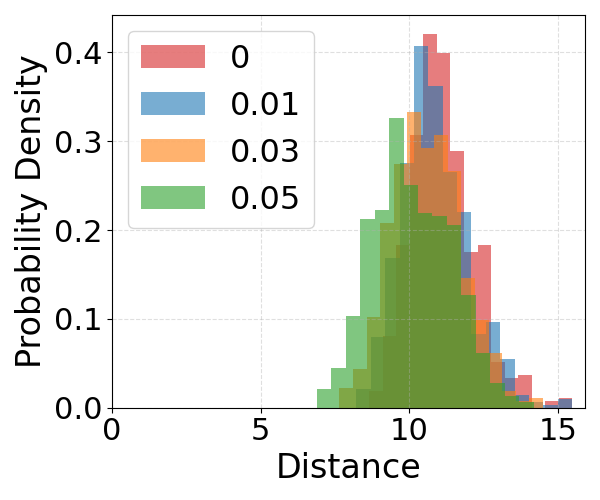}
        \caption{}
        \label{fig:thin_shell}
    \end{subfigure}

    \vspace{0.3cm}

    % ---- Second Row ----
    \begin{subfigure}[t]{0.42\linewidth}
        \centering
        \includegraphics[width=\linewidth]{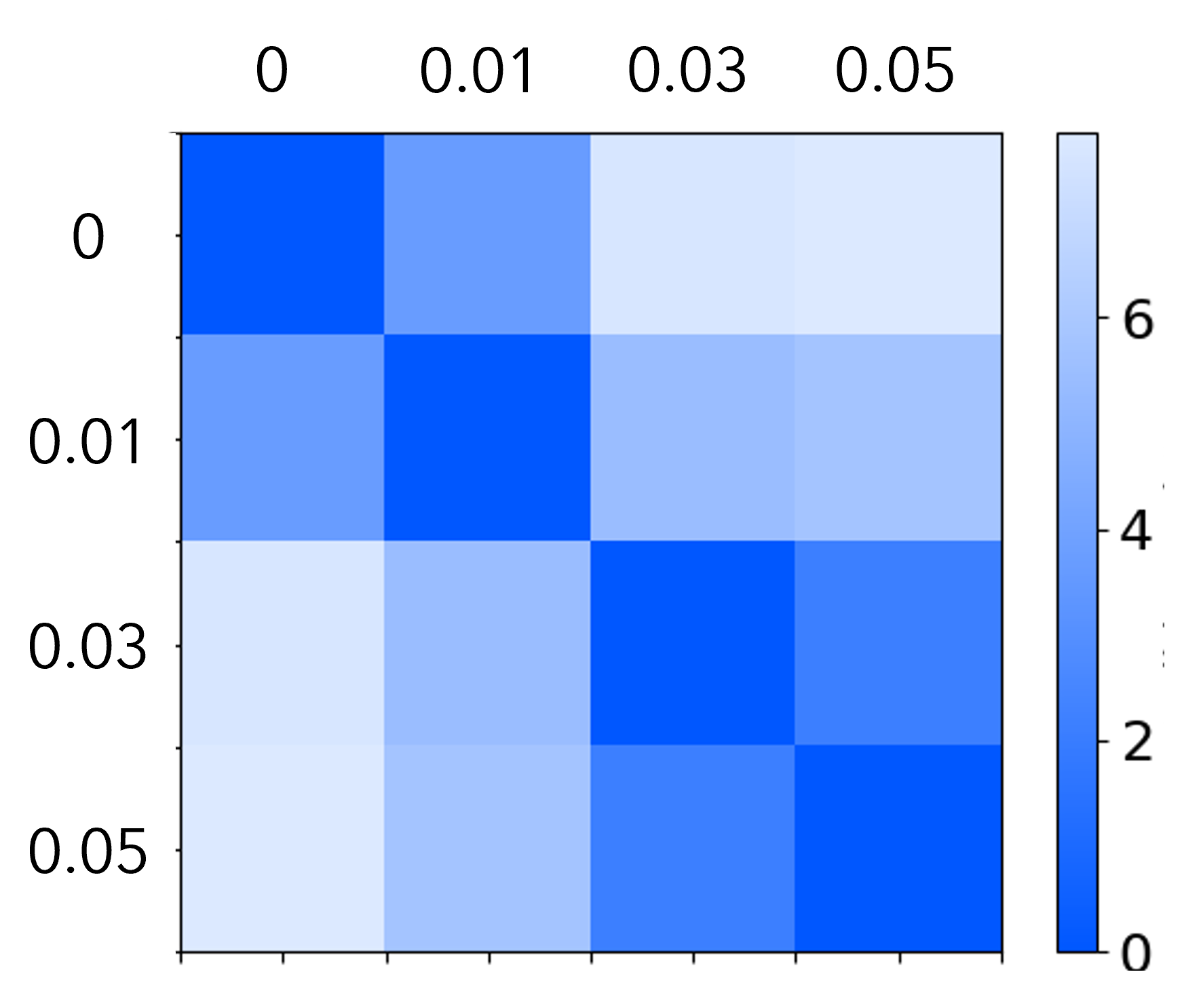}
        \caption{}
        \label{fig:heat_map}
    \end{subfigure}
    \hfill
    \begin{subfigure}[t]{0.42\linewidth}
        \centering
        \includegraphics[width=\linewidth]{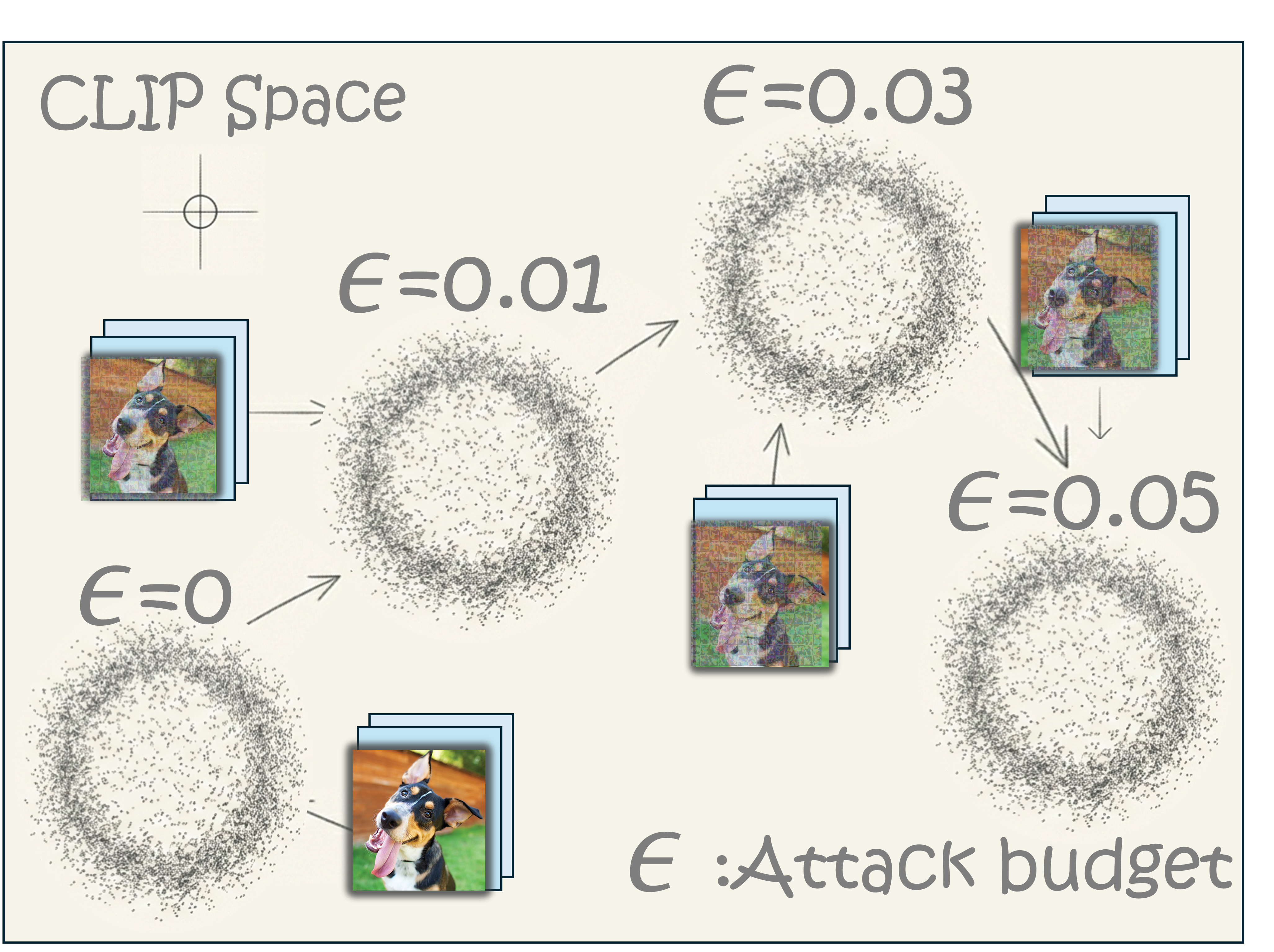}
        \caption{}
        \label{fig:illustration}
    \end{subfigure}

    \caption{\textbf{CLIP sensitivity to adversarial perturbations.} 
    \subref{fig:cos_sim} Cosine similarity histogram between Tiny-ImageNet images and their FGSM adversarial counterparts, showing a clear embedding shift even for small budgets. 
    \subref{fig:thin_shell} \textbf{Thin-shell structure.} Histogram of embedding norms with respect to the dataset mean reveals that samples lie on a thin shell~\cite{levi2025double} (much larger bias than variance). 
    \subref{fig:heat_map} \textbf{Gradual latent shift.} As the attack budget increases, the $\ell_2$ distance between distribution centers grows consistently. 
    \subref{fig:illustration} \textbf{Embedding evolution under attack.} Increasing perturbation strength progressively shifts samples in the latent space.}
    \label{fig:embedding_geometry}
\end{figure}

Our motivation is based on two key observations regarding the behavior of CLIP under adversarial perturbations. First, although adversarial changes are often visually imperceptible, CLIP embeddings are sensitive to such perturbations (previous works on ImageNet-C \cite{hendrycks2021natural} shows that CLIP norms gradually grow as noise increases \cite{betser2025whitened, levi2025double}). Even small attacks induce a measurable shift in the representation, reflected by a non-negligible change in cosine similarity between clean and attacked images (see Fig.\ref{fig:cos_sim}). Second, this shift is not arbitrary and yields a distinct bias, which can be well measured
(see Fig.\ref{fig:thin_shell}, Fig. \ref{fig:heat_map} and Fig. \ref{fig:illustration}).
As attacked samples concentrate in a specific region of the latent space, we can leverage the semantic structure of CLIP to characterize this region using text prompts, and identify adversarial inputs based on their similarity to it.

Based on these observations, we propose \textit{$A^4D$ (\textbf{A}ttack- and \textbf{A}rchitecture-\textbf{A}gnostic \textbf{A}dversarial \textbf{D}etector)}, a zero-shot method that operates in the embedding space of CLIP. By computing the cosine similarity between an image and a small set of dedicated captions, $A^4D$ can detect adversarial inputs without access to the classifier or knowledge of the attack.

Our main contributions are:
\begin{itemize}
\renewcommand{\labelitemi}{$\bullet$}
\item We propose to utilize CLIP as \textbf{a zero-shot adversarial detection framework}, based on the similarity to a predefined set of prompts.
\item The method is \textbf{attack- and classifier-agnostic}, with limited validation-based design.

\item We demonstrate the effectiveness of $A^4D$ on multiple attacks (FGSM, PGD, AutoAttack, BIM, CW, DeepFool and Square), two Datasets (Tiny-ImageNet, StreetSurfaceVis) and four classifiers (DeiT-Small, Wide-ResNet, ResNet34 and convnext-tiny), yielding SOTA performance.

\item The framework is 
\textbf{computationally efficient}: training requires a single forward pass to compute embeddings, and detection at test time involves lightweight operations on low-dimensional similarity vectors.
\end{itemize}

% %%%%%%%%%%%%%%%%%%%%%%%%%%%%%%%%%%%%%%%%%%

% %%%%%%%%%%%%%%%%%%%%%%%%%%%%%%%%%%%%%%%%%%

\begin{figure}[tb]
    \centering
    \includegraphics[height=6.2cm]{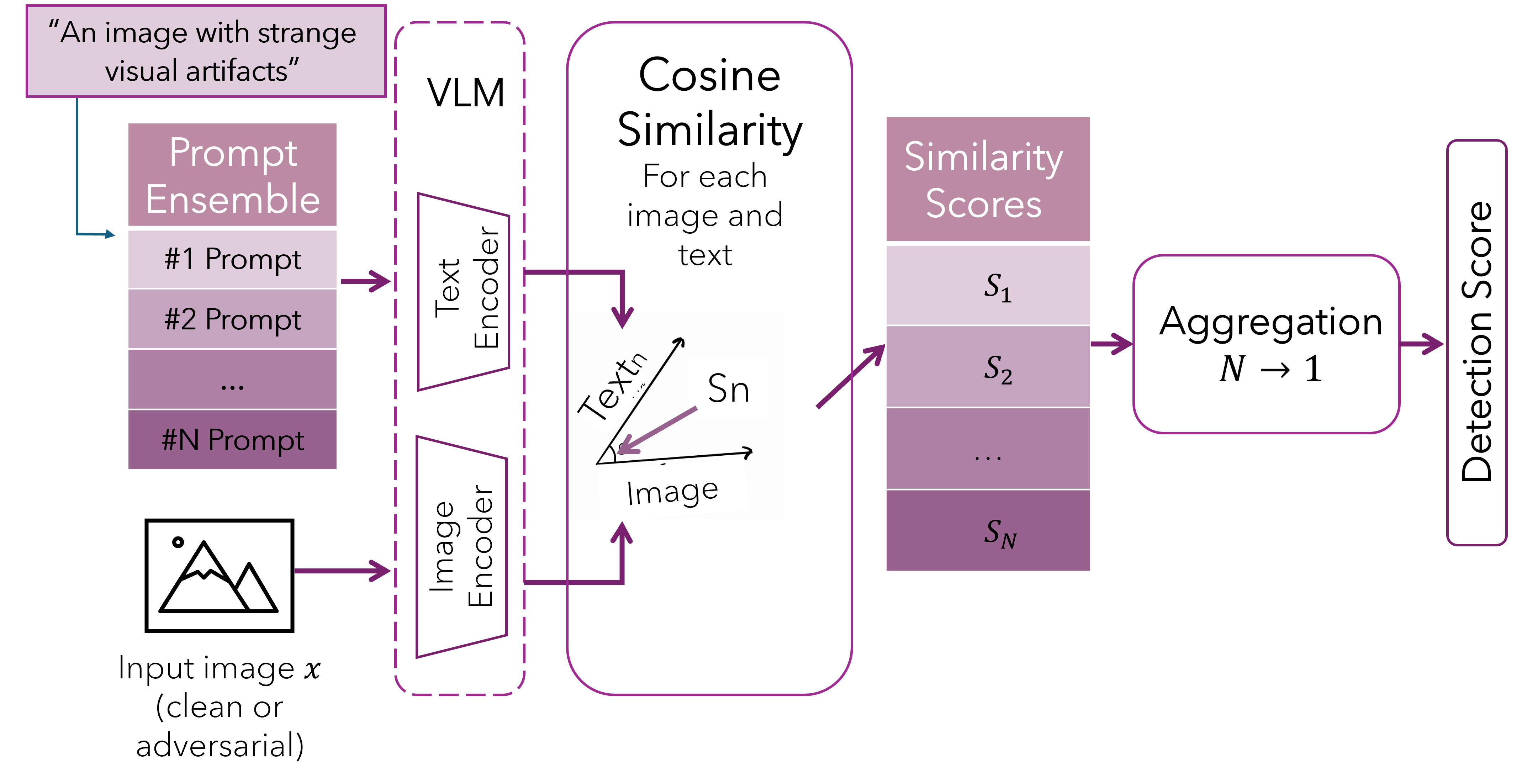}
    \caption{\textbf{Test-time zero-shot adversarial detection pipeline}. Given an input image, cosine similarities between the image embedding and a fixed set of textual prompt embeddings are computed using CLIP. The resulting similarity scores are aggregated into a single detection score, where higher values indicate a higher likelihood of adversarial perturbation.}
    \label{fig:method}
\end{figure}

\section{Related Work}
\label{sec:related}
\subsection{Detection of Adversarial Attacks}
 
Broadly, adversarial detection methods can be grouped into three categories:

\textbf{1. Attack-specific detection methods}. These methods are designed to detect a particular attack or a limited family of attacks. Typical algorithms include learning a threshold or training a classifier to distinguish between clean and adversarial samples using confidence scores, input statistics, or output distributions \cite{metzen2017detecting, feinman2017detecting, xu2017feature}.
While this direction can be effective under the assumed threat model, it often relies on attack-dependent parameters. This implicitly assumes prior knowledge of the attack type, and in some cases even of the attack parameters. This type of information is often not available for defense in real-world scenarios.

\textbf{2. Internal classifier representations.} These techniques use feature activations or hidden layers, to detect adversarial inputs \cite{ma2018characterizing, papernot2018deep}, or exploit architecture-specific characteristics (e.g., properties of convolutional neural networks) \cite{nazeri2025entropy}. An inherent underlying assumption is the use of a particular model structure, which may not hold for all classifiers. In these approaches, the detection algorithm is tightly coupled to the specific classifier architecture.

\textbf{3. Training-based detection.} These approaches rely on training a new model, either an explicit detector or a modified version of the classifier itself. In these methods, detection is often formulated as a supervised learning problem, where the model is trained using both clean and adversarial examples, e.g., \cite{metzen2017detecting, meng2017magnet}. Such approaches usually require access to attacked data during training, and their performance strongly depends on the attacks used to generate this data.

In contrast, we focus on a zero-shot detection setting that is agnostic to both the attack type and the underlying classifier. The proposed formulation does not require labeled adversarial data to characterize the distinction between clean and adversarial inputs. To the best of our knowledge, this highly challenging setting  has not been addressed in prior work.

\subsection{Usages of CLIP}
CLIP \cite{radford2021learning} is a pioneering multi-modal encoder trained to align images and text into a shared embedding space using contrastive learning on large-scale image–text pairs.
Beyond classification, CLIP has been applied to a variety of tasks, including retrieval \cite{sultan2023exploring, luo2022clip4clip}, open-vocabulary recognition \cite{weng2023open, wu2023clipself}, and text2image \cite{ramesh2021zero, saharia2022photorealistic}. More recently, CLIP-based representations have been explored for out-of-distribution and anomaly detection in a zero-shot setting. For example, AnomalyCLIP \cite{zhou2023anomalyclip} uses textual prompts to detect anomalous regions without task-specific training, AdaCLIP \cite{cao2024adaclip} adapts CLIP representations for anomaly detection across domains, and AA-CLIP \cite{ma2025aa} leverages CLIP embeddings to identify abnormal samples without access to anomaly labels. In this work we show that CLIP can be highly instrumental also in the adversarial attack detection field.

\begin{figure*}[tb]
    \centering
    \includegraphics[width=1\linewidth]{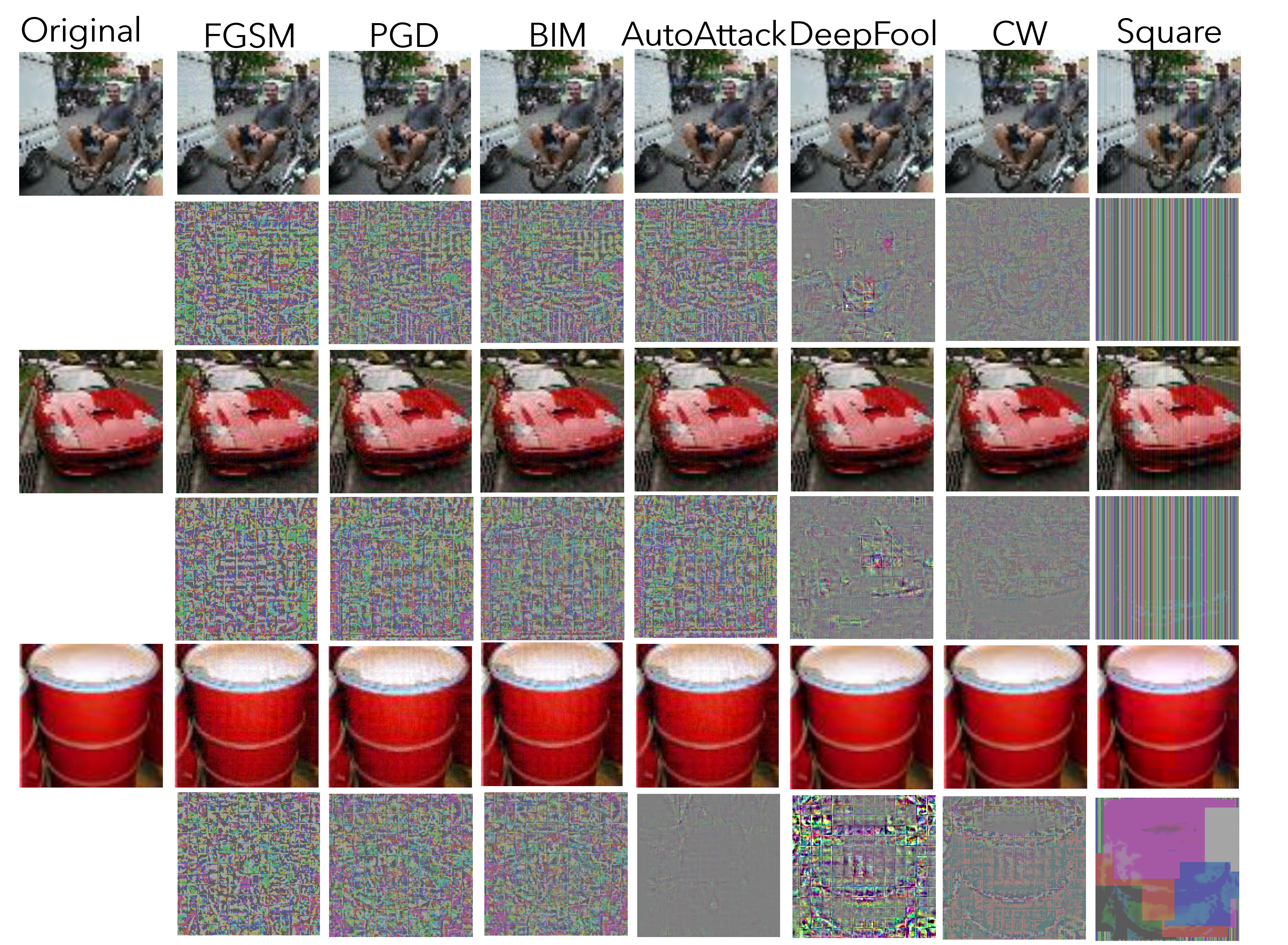}
    \caption{\textbf{Qualitative examples of clean and adversarial images under different attack methods.} Three pairs of clean and adversarial images are shown. The second row in each example visualizes the difference between the adversarial and original images (with scale). All attacks are applied on the DeiT-Small classifier.}
    \label{fig:all_attacks}
\end{figure*}

\begin{figure}[ptbh]
    \centering
    \includegraphics[width=0.9\linewidth]{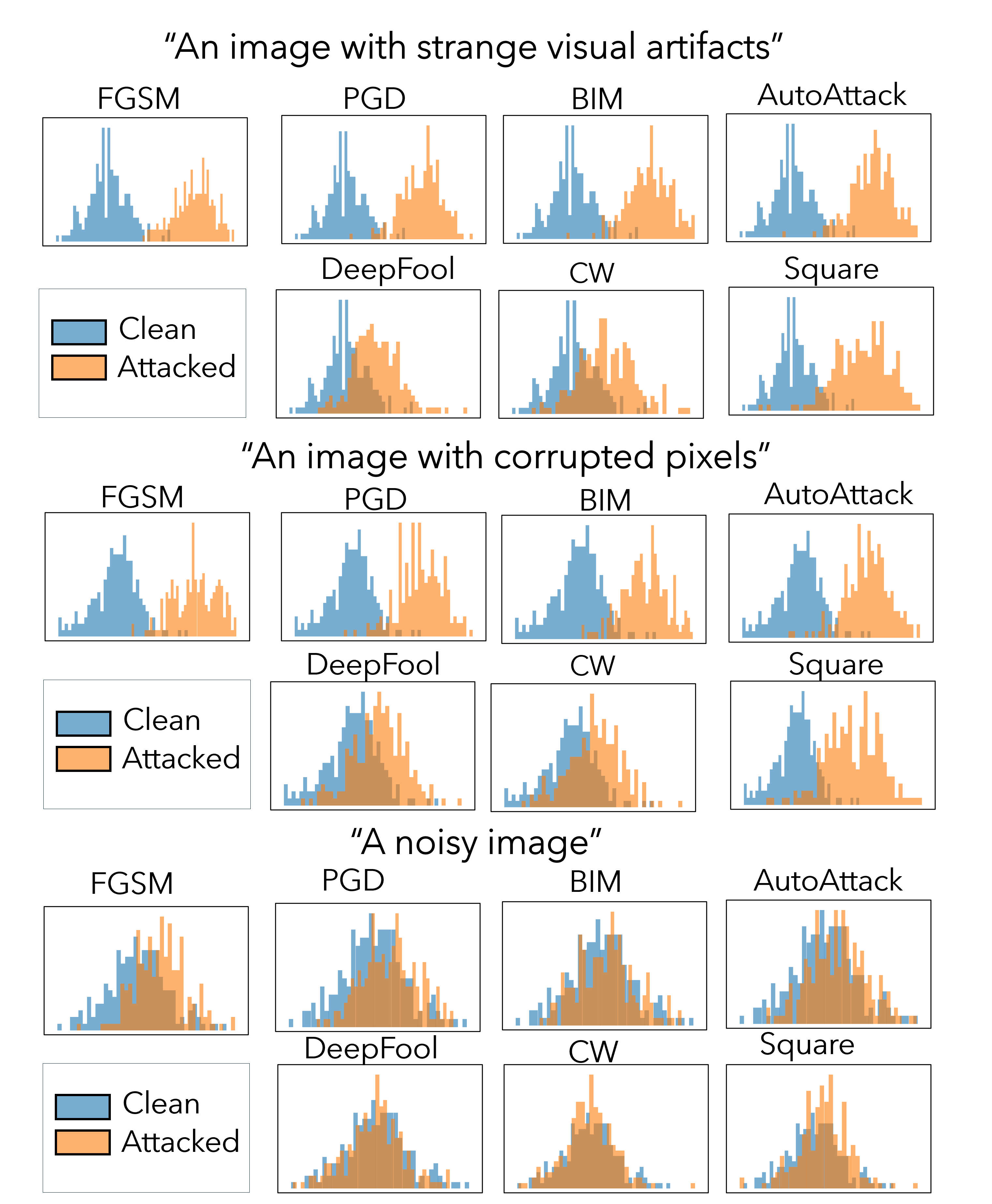}
    \caption{
    \textbf{Prompt-specific detection behavior.}
    Three examples illustrating the similarity between images and three specific text prompts are shown. Only the first two prompts are included in the final prompt dictionary.
    For the first two prompts, a clear separation between clean and adversarial samples is observed for most attack types, with DeepFool and CW being the most challenging to distinguish. The behavior varies across prompts, as each prompt exhibits different sensitivity patterns to specific attacks.
    Since our goal is to maintain a zero-shot setting with the potential to generalize to previously unseen attacks, we employ multiple prompts in the final method to enhance robustness and diversity.
    The third prompt serves as a counterexample. Although it is semantically similar to the first two, it yields poor separation when used as a caption, highlighting the importance of careful prompt selection.
    }
    \label{fig:prompt_examples}
\end{figure}

\section{Method}
\label{sec:method}
\subsection{Preliminaries and problem definition
}
Let $X \in \mathbb{R}^{H \times W \times C}$ denote an image and let
$t \in \mathcal{T}$ denote a text prompt.  
The corresponding CLIP embeddings in the shared embedding space are denoted by 
$z_I \in \mathbb{R}^d$ for an image and $z_T \in \mathbb{R}^d$ for a text.
The cosine similarity between the image and text embeddings is denoted by $sim(z_I, z_T) = \frac{\langle z_I,z_T \rangle}{\|z_I\|\|z_T\|}$. We denote vectors by lowercase letters and matrices by uppercase letters.
A classifier $c : \mathbb{R}^{H \times W \times C} \rightarrow \{1, \dots, K\}$ maps an input image to one of $K$ class labels. An adversarial example $X^{adv} = X + \delta$ is generated using an untargeted attack, 
where the perturbation is limited to a budget $\epsilon$, such that $\|\delta\|_p \leq \epsilon$ , with $p$ corresponding to the standard norm for each attack (e.g., $\ell_\infty$ for FGSM, PGD, BIM, Square; $\ell_2$ for DeepFool and CW), and aims to cause misclassification, i.e.,
$c(X^{adv}) \neq c(X)$.
A detector 
$d : \mathbb{R}^{H \times W \times C} \rightarrow \{0,1\}$ is tasked with correctly predicting whether an input image $X'$ is clean or adversarial, that is, 
$d(X') = 0$ if $X'$ is clean and $d(X') = 1$ if $X'$ is adversarial.

\subsection{Exploiting structured latent behavior for adversarial detection}

As shown in Fig.~\ref{fig:illustration}, CLIP embeddings are sensitive even to small, imperceptible perturbations in the image domain, such that $z^{adv}_I \neq z^{clean}_I$, and embedding distributions $z^{adv}_I$ and $z^{clean}_I$ shift as the attack budget grows. This raises the question: \textit{How can we leverage this latent behavior to detect malicious samples?}  
A simple, naïve approach would be to classify images based on their distance from the expected endpoint of this trajectory (full details are in the Supp.). As shown in Table~\ref{tab:dist_to_noise}, while this detector achieves reasonable performance, it remains limited due to several reasons: (1) it assumes that all attacks produce similar shifts, whereas we aim for an attack-agnostic detector; (2) it ignores the semantic knowledge of CLIP, since it does not involve the text domain, which could provide a more interpretable solution.  
To address these limitations, we propose projecting images onto a dictionary of related captions. This approach leverages CLIP’s semantic space, reduces overfitting to specific attacks, and enhances robustness to a variety of attack mechanisms while providing a more explainable detection signal.

\begin{table*}[tb]
\centering
\caption{\textbf{Na\"ive detector.} Zero-shot adversarial detection performance by measurement the distance to trajectory endpoint (AU-ROC) on Tiny-ImageNet dataset.}
\label{tab:dist_to_noise}
\begin{tabular}{lccccccc}
\toprule
Method & FGSM & PGD & AutoAttack & BIM & CW & DeepFool & Square \\
\midrule
Distance to endpoint & 93.87 & 89.21 & 87.79 & 82.37 & 66.31 & 66.78 & 85.36\\
\bottomrule
\end{tabular}
\end{table*}

\subsection{Our adversarial attack detection framework ($A^4D$)}
\begin{figure}[tb]
    \centering
    \includegraphics[width=0.5\linewidth]{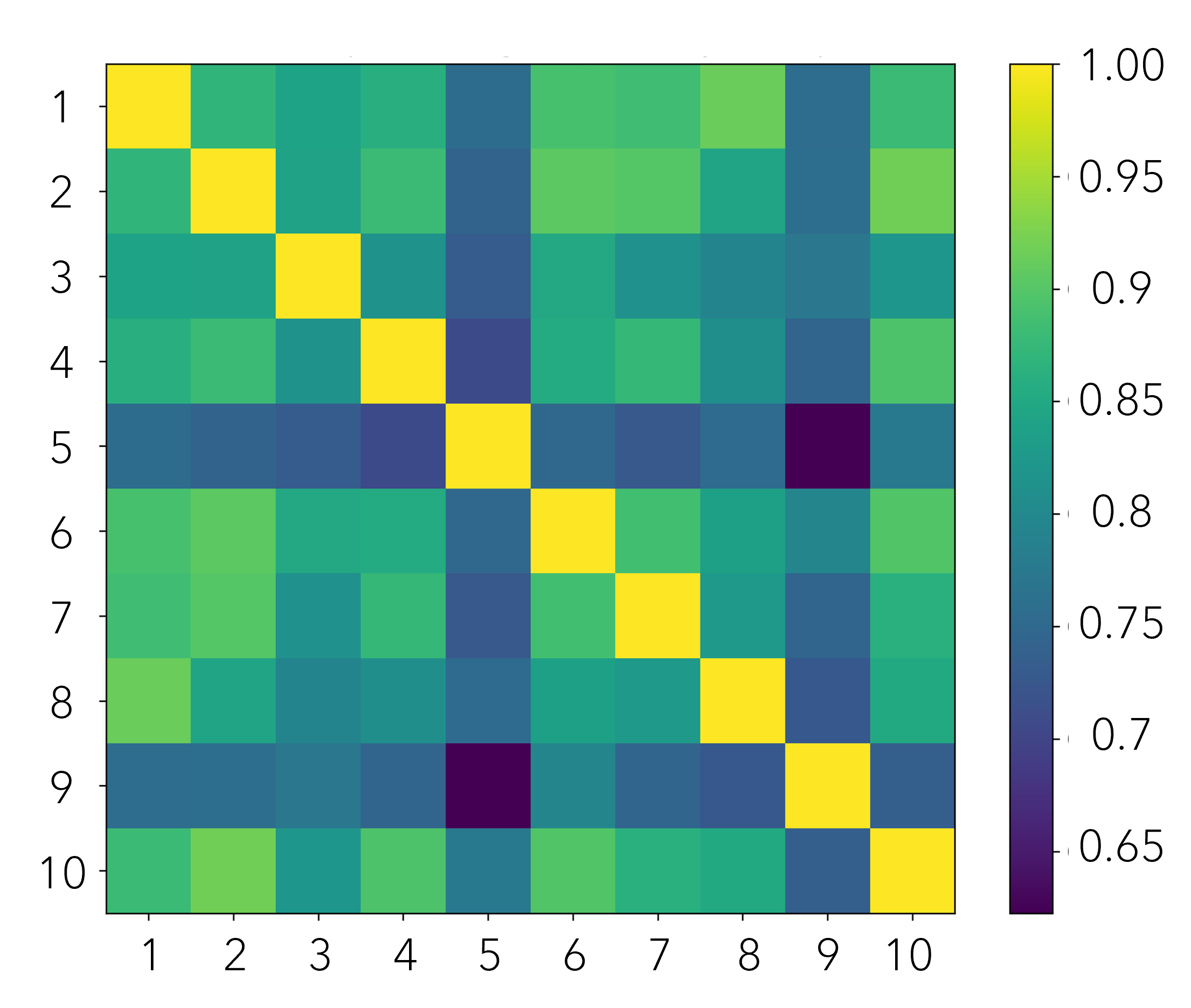}
    \caption{Pairwise correlation heatmap of the prompt-based similarity scores computed using CLIP. Prompts are indexed according to 
    %Appendix~A. 
    Table~\ref{tab:prompts}.
    The low inter-prompt correlations suggest that the selected prompts provide diverse indicators.}
    \label{fig:clip_prompts}
\end{figure}

First, we construct a dictionary of text prompts that describe the semantic changes induced by noise or adversarial attacks. Leveraging CLIP’s strong semantic understanding, we select $N$ text prompts that capture different forms of perturbations, shown in Table~\ref{tab:prompts}.
\begin{table}[t]
\centering
\caption{The dictionary of text prompts used for detecting adversarial images ($N = 10$).}
\begin{tabular}{cl}
\hline
\textbf{ID} & \textbf{Prompt} \\
\hline
1 & ``An image with noise'' \\
2 & ``An image with unnatural texture'' \\
3 & ``An image with digital distortions'' \\
4 & ``An image with corrupted pixels'' \\
5 & ``An image with adversarial perturbations'' \\
6 & ``An image that was modified artificially'' \\
7 & ``An image with strange visual artifacts'' \\
8 & ``An image with high-frequency noise'' \\
9 & ``An image that contains tampering'' \\
10 & ``An image with pixel-level manipulation'' \\
\hline
\end{tabular}
\label{tab:prompts}
\end{table}
We employ an ensemble of prompts for two main reasons: (i) to capture variations across multiple attack types, since the embedding-space behavior of different attacks may not be identical (see Fig. 
\ref{fig:prompt_examples}), and (ii) to reduce variance in the detection signal, thereby stabilizing the algorithm (see the low correlation between prompts in Fig. \ref{fig:clip_prompts}).

Second, we empirically estimate the distribution of cosine similarities between clean images and the text prompt dictionary. To do this, we construct a representative set of $M$ clean images from the dataset of interest. For each image $X_i$ and each prompt $t_j$, $j = 1,\dots,N$, we compute the similarity $sim(z^i_I, z^j_T)$. %$sim(z^i_i, z^j_t)$. 
This allows us to characterize the “similarity space” spanned by clean images, an $N$-dimensional space where each axis corresponds to the similarity with a specific prompt. Later, this space is used to determine whether a new image $X'$ lies within the expected distribution of clean images or originates from a different distribution (e.g., adversarially perturbed).

Finally, we should aggregate the $N$-dimensional similarity vector to assign a single score, $s \in \mathbb{R}$, to each image. 
We compare several reasonable aggregation methods (see Table \ref{tab:results_Tiny_minmax}) and find that the PCA-based aggregation performs best. Therefore, we describe this method in greater detail in the next section.

\subsection{Detecting adversarial attacks at inference}

Let $Z_I \in \mathbb{R}^{M \times d}$ be the embedding matrix of the representative set, 
where $M$ is the number of clean images and $d$ is the embedding dimension. 
This representative set serves as a small validation set used solely for estimating the aggregation direction and does not require model training or fine-tuning. In practice, we use only 200 clean images sampled from the target dataset, which is sufficient to obtain a stable estimate of the principal component.
Let $Z_T \in \mathbb{R}^{N \times d}$ be the embedding matrix of the dictionary 
of text prompts, where $N$ is the number of prompts.
We define $P \in \mathbb{R}^{M \times N}$ such that each entry  $P_{ij}= 
sim(z^{(i)}_I, z^{(j)}_T)$.

The matrix $P$ undergoes standardization, yielding $\tilde{P}$. Namely, each column of $\tilde{P}$ has mean $0$ and standard deviation $1$. Then, PCA is applied to $\tilde{P}$, and $v_1$, the direction of the first principal component, is obtained.

Given a new image $X' \in \mathbb{R}^{H \times W \times C}$, we obtain its embedding $z'_I$ and compute its similarity to dictionary prompt embeddings $u'_j = \mathrm{sim}(z'_I, z_T^{(j)})$, $j=1,\dots,N$, yielding $u' \in \mathbb{R}^N$. The previously calculated standardization transformation is applied to $u'$, yielding $\tilde{u}'$. The detection score is obtained by projecting $\tilde{u}'$ onto the first principal direction $v_1 \in \mathbb{R}^N$:
\begin{equation}
s' = \tilde{u}'^\top v_1. 
\end{equation}

%================================================

The use of the direction of PC1 for aggregation is motivated by the fact that it is the one-dimensional linear projection that explains the maximum variance in the similarities across instances in the representative set.

The score $s'$ is used to compute the AUC on the test set. The overall pipeline of the method is illustrated in Fig.~\ref{fig:method}.

\section{Experiments}
\label{sec:exp}
\subsection{Datasets}
We evaluated our method on (1) Tiny-ImageNet \cite{tiny-imagenet} contains 200 classes. In addition to this benchmark we validate the results on a smaller dataset (2) StreetSurfaceVis \cite{kapp_streetsurfacevis_2025}, which contains 5 classes of different road materials. The non-standard dataset was used as additional verification of our method since it was published after the introduction of CLIP. The second dataset allows us to validate that the results are not linked to any possible exposure of CLIP during training with the standard dataset (which was published earlier). 

% \subsection{Models}
\subsection{Adversarial Attacks}
We used torchattacks \cite{kim2020torchattacks} to implement all examined attacks; They are mostly white-box: Fast Gradient Sign Method (FGSM) \cite{goodfellow2014explaining}, Projected Gradient Descent (PGD) \cite{madry2017towards}, Basic Iterative Method (BIM) \cite{kurakin2018adversarial}, AutoAttack \cite{croce2020reliable}, DeepFool \cite{moosavi2016deepfool}, Carlini-Wagner (CW) \cite{carlini2017towards} and one black-box attack - Square Attack \cite{andriushchenko2020square}.
Examples of the attacks are shown in Fig.~\ref{fig:all_attacks}. Pay attention that while some attacks are noise-like, there exists attacks which their additive signal is more structured.

For all attacks that require a perturbation budget (FGSM, PGD, BIM, AutoAttack, and Square), we set the budget to $\epsilon = 8/255$. For iterative attacks (PGD and BIM), we use a fixed number of iterations with step size ($\alpha_{PGD} = \epsilon/4, \alpha_{BIM} = \epsilon/4$), chosen according to standard practice. AutoAttack and Square are evaluated using their official implementations with default parameters, except for the perturbation budget $\epsilon = 8/255$. All remaining attacks are configured following their original papers based on their official implementations.

\begin{figure}[tb]
    \centering
    \includegraphics[width=\linewidth]{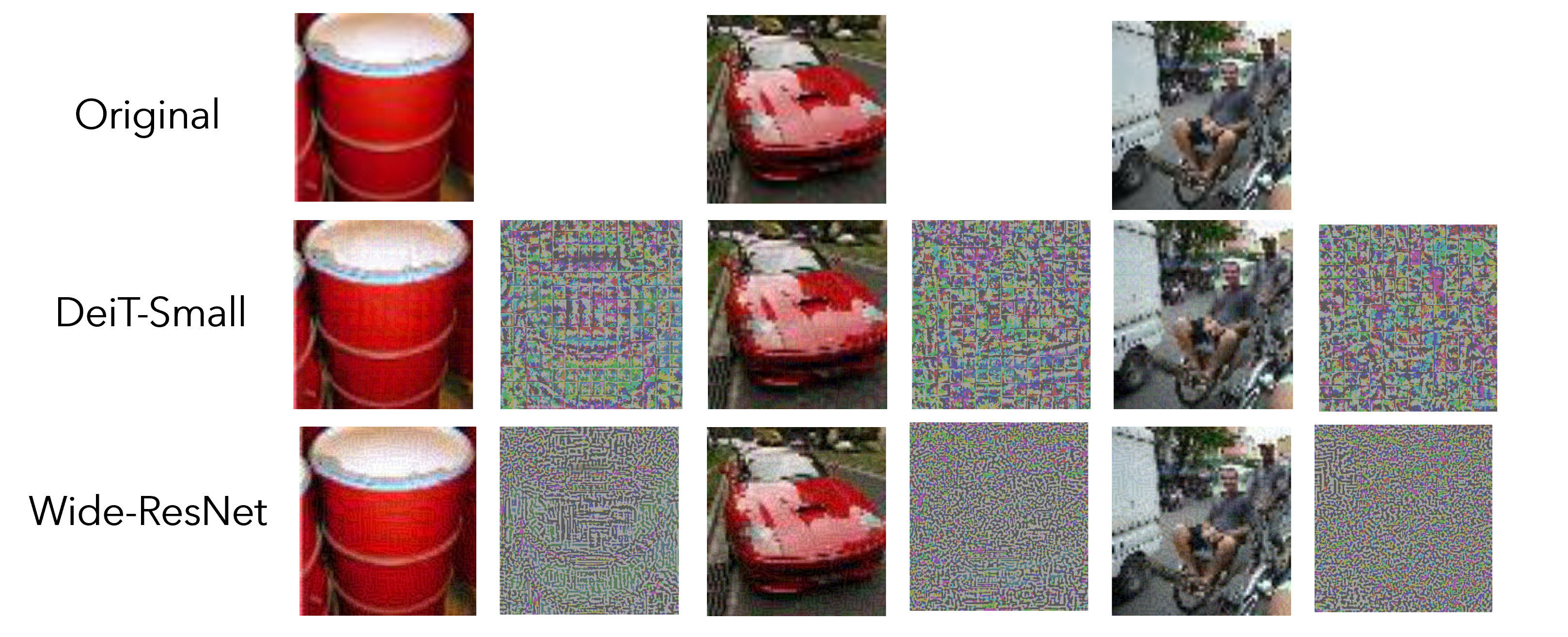}
    \caption{\textbf{Qualitative examples of clean and adversarial images under the same attack (FGSM) across different models}. Three pairs of clean and adversarial images are shown. The image to the right of each clean image shows the difference between the attacked image and the clean image (with increased contrast). The second row corresponds to the DeiT-Small architecture, and the third row corresponds to the Wide-ResNet architecture. The difference images clearly show that the same attack (FGSM in this example) produces different perturbation patterns for the two classifiers.}
    \label{fig:model_agnostic}
\end{figure}

\subsection{Experimental Setup}
We evaluate the proposed method on two image datasets and four classifier architectures: Wide-ResNet \cite{zagoruyko2016wide}, ResNet34 \cite{he2016deep}, DeiT-Small \cite{touvron2021training} and ConvNext-Tiny \cite{liu2022convnet}. The use of multiple classifiers demonstrates the classifier-agnostic nature of the method. Since each classifier responds differently to adversarial optimization, the resulting perturbations produced by the same attack may vary across models, as illustrated in Figure~\ref{fig:model_agnostic}.

\subsubsection{Runtime.} Since our method is almost training-free (requires only mini validation set), we only report inference time.
All experiments are conducted on a single NVIDIA RTX~3090 GPU with batch size 64.
The average inference time is 0.0091 seconds per image, measured over 200 images and excluding data loading.
The reported runtime includes evaluation over all textual prompts used by our method. This runtime enables real-time processing and large-scale evaluation.

\subsection{Baselines}
For comparison, we implement two classical classifier-dependent baselines based on noise statistics: Median Absolute Deviation (MAD) and  Wavelet-based noise estimation. In addition, we re-implement a prior method for distinguishing clean and adversarial examples based on  Mahalanobis \cite{lee2018simple}, which was originally proposed as a generalized detection approach. To assess this claim, we evaluate two variants of this method: (i) an \textit{attack-aware} version that is tuned separately for each attack, and (ii) an \textit{attack-agnostic} version that does not assume knowledge of the attack type.

Due to the high computational cost of the baseline methods, these comparisons are not performed on AutoAttack of two classifier architectures. This setting follows common practice and enables a fair comparison under feasible computational constraints.

One of the evaluated classifier architectures (ResNet34) exhibits lower classification accuracy compared to the others on the Tiny-ImageNet dataset. The Mahalanobis baseline was crafted for a set of architectures including ResNet34; therefore, to enable a faithful comparison with the Mahalanobis baseline, this model was  included. Following the original experimental setup allows us to minimize deviations from prior work and isolate differences which can be attributed to the detection method rather than to architectural choices.

\begin{figure*}[tb]
    \centering
    \includegraphics[width=0.8\linewidth]{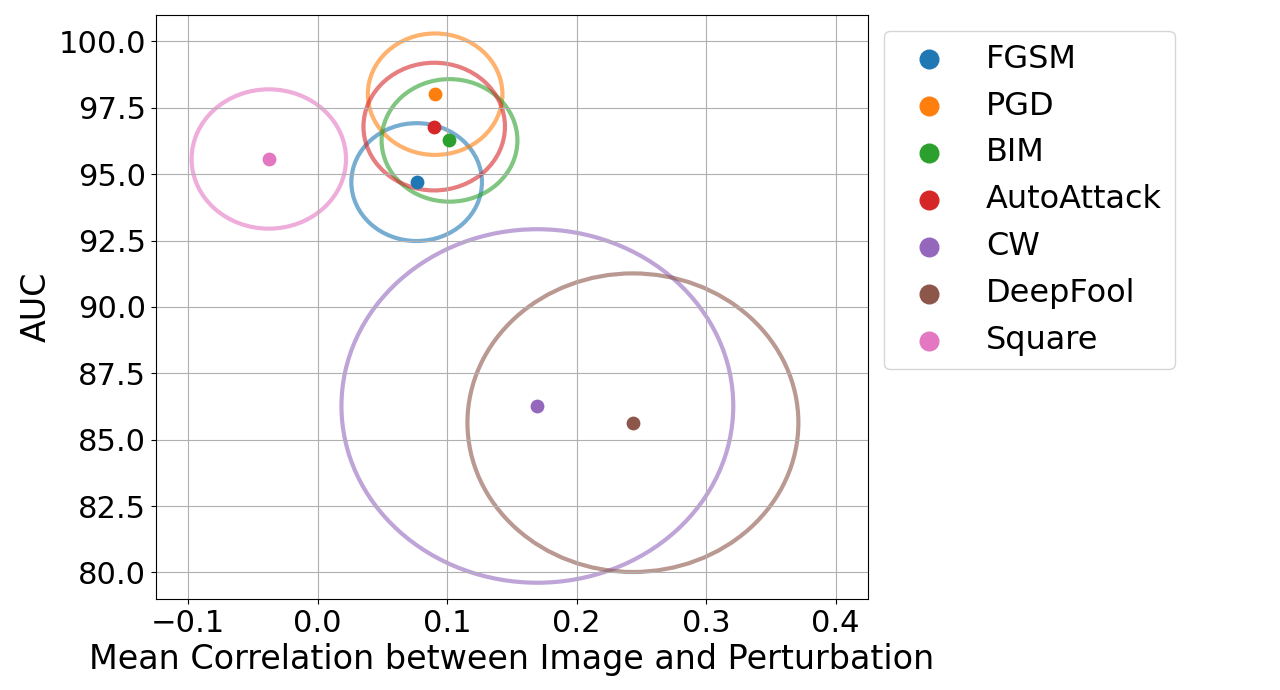}
    \caption{\textbf{Relationship between perturbation characteristics and detection performance.}
    Each point represents a different adversarial attack. The x-axis denotes the mean correlation between the original image and the corresponding difference image (original minus adversarial), measuring the extent to which the perturbation preserves image structure. The y-axis shows the AUC of our detection method (tiny-ImageNet Dataset; DeiT-Small classifier). The radius of each circle corresponds to the standard-deviation of the correlations.}
    \label{fig:corr_auc}
\end{figure*}
\section{Evaluation}

\subsection{Results}
The main results of the proposed method across all datasets and classifier architectures, along with a detailed comparison with baseline methods is presented in Table~\ref{tab:results_comp}, where results corresponding to the attack-agnostic variant of the Mahalanobis method are marked with (*). Beside on ConvNext-Tiny, where our approach performs only second-best on average, on all other architectures our method is optimal or second best on all attacks, and best on their average.

\begin{table*}[tb]
\centering
\footnotesize
\caption{\textbf{Zero-shot adversarial detection performance (AU-ROC) for multiple architectures and attack methods.}
The competitors (MAD, Wavelet) are methods that work only under architecture- and attack-agnostic setting. For comparison, we add (in gray shading) two variants of Mahalanobis method \cite{lee2018simple}: the regular and a class agnostic variant (marked with *). \textbf{Bold} - best, \underline{underline} - second best.
}
\label{tab:results_comp}
\begin{tabular}{p{1cm} lccccccc|c}
\toprule
Model & Method & FGSM & PGD & \shortstack{Auto\\Attack} & BIM & CW & \shortstack{Deep\\Fool} & Square & Avg. \\
\midrule
\multirow{5}{*}{\shortstack{Wide \\ResNet}}
& $A^4D$ \textbf{(Ours)} & \textbf{99.64} & \underline{99.67} & \textbf{99.75} & \underline{99.44} & \textbf{78.03} & \textbf{77.55} & \textbf{93.61} & \textbf{92.53}\\
& MAD & \underline{85.80} & 79.08 & 82.36 & 62.94 & 50.02 & 49.98 & \underline{81.45} & 70.23\\
& Wavelet & \textbf{99.64} & 97.43 & 97.68 & 82.88 & 50.48 & 50.11 & 70.28 & \underline{78.36}\\
& \cellcolor{gray!30}Mahalanobis & \cellcolor{gray!30}67.10 & \cellcolor{gray!30}\textbf{99.91} & \cellcolor{gray!30}\underline{99.24} & \cellcolor{gray!30}\textbf{99.76} & \cellcolor{gray!30}\underline{55.26} & \cellcolor{gray!30}\underline{55.44} & \cellcolor{gray!30}62.08 &\cellcolor{gray!30}76.97 \\
& \cellcolor{gray!30}Mahalanobis* & \cellcolor{gray!30}67.10 &\cellcolor{gray!30}82.91 &\cellcolor{gray!30}81.60 &\cellcolor{gray!30}78.64 &\cellcolor{gray!30}54.65 &\cellcolor{gray!30}55.10 &\cellcolor{gray!30}50.02 &\cellcolor{gray!30}67.15\\
\hline
\multirow{5}{*}{\shortstack{Res-\\Net34}}
& $A^4D$ \textbf{(Ours)} & \underline{94.70} & \textbf{98.01} & \underline{96.79} & \textbf{96.27} & \textbf{86.27} & \textbf{85.64} & \textbf{95.57} & \textbf{93.32}\\
& MAD & 85.55 & 78.85 & 82.20 & 63.34 & 50.17 & 50.00 & \underline{82.94} & 70.43 \\
& Wavelet & \textbf{99.58} & \underline{97.42} & \textbf{97.63} & \underline{82.49} & 50.46 & 50.02 & 70.46 & 78.29\\
& \cellcolor{gray!30}Mahalanobis & \cellcolor{gray!30}91.94 &\cellcolor{gray!30}91.52 & \cellcolor{gray!30}85.05 & \cellcolor{gray!30}77.27 &\cellcolor{gray!30}\underline{70.99} &\cellcolor{gray!30}\underline{68.58} &\cellcolor{gray!30}67.22 &\cellcolor{gray!30}\underline{78.94} \\
& \cellcolor{gray!30}Mahalanobis* & \cellcolor{gray!30}91.94 & \cellcolor{gray!30}13.14 & \cellcolor{gray!30}29.33 & \cellcolor{gray!30}18.59 & \cellcolor{gray!30}38.57 & \cellcolor{gray!30}38.24 & \cellcolor{gray!30}42.61 & \cellcolor{gray!30}38.92 \\
\hline
\multirow{5}{*}{\shortstack{Deit-\\Small}}
& $A^4D$ \textbf{(Ours)} & \textbf{99.24} & \textbf{98.16} & \textbf{98.39} & \textbf{97.76} & \textbf{77.31} & \textbf{78.62} & \textbf{97.84} & \textbf{92.47}\\
& MAD & 73.46 & 71.07 & 85.95 & 83.46 & 50.68 & 50.46 & 63.52 & 68.37\\
& Wavelet & \underline{95.99} & \underline{95.05} & \underline{93.69} & \underline{90.36} & 50.92 & 50.62 & 56.88 & \underline{76.22}\\
& \cellcolor{gray!30}Mahalanobis & \cellcolor{gray!30}61.45 & \cellcolor{gray!30}61.94 & \cellcolor{gray!30}-  &\cellcolor{gray!30}62.53 & \cellcolor{gray!30}\underline{63.82} &\cellcolor{gray!30}\underline{63.55} & \cellcolor{gray!30}\underline{69.08} & \cellcolor{gray!30}63.72\\
& \cellcolor{gray!30}Mahalanobis* & \cellcolor{gray!30}61.45 &\cellcolor{gray!30}60.59 & \cellcolor{gray!30}- & \cellcolor{gray!30}60.51 & \cellcolor{gray!30}47.06 & \cellcolor{gray!30}48.27 &\cellcolor{gray!30}51.92 &\cellcolor{gray!30}54.96\\
\hline
\multirow{5}{*}{\shortstack{Conv-\\next\\ tiny}}
& $A^4D$ \textbf{(Ours)} & 93.80 & 95.38 &  \textbf{97.87} & 83.18 & \underline{62.83} & \underline{63.47} & \textbf{98.24} & \underline{84.97}\\
& MAD & 77.73 & 69.64 & 72.29 & 57.12 & 50.10 & 50.01 & 76.65 & 64.79 \\
& Wavelet & \underline{94.07} & 87.33 & \underline{87.68} & 69.88 & 50.39 & 50.24 & 70.44 & 72.86 \\
& \cellcolor{gray!30}Mahalanobis & \cellcolor{gray!30}\textbf{96.46} & \cellcolor{gray!30}\underline{97.28} & \cellcolor{gray!30}- & \cellcolor{gray!30}\underline{97.65} & \cellcolor{gray!30}\textbf{74.64} & \cellcolor{gray!30}\textbf{75.13} & \cellcolor{gray!30}\underline{91.98} & \cellcolor{gray!30}\textbf{88.85} \\
& \cellcolor{gray!30}Mahalanobis* & \cellcolor{gray!30}\textbf{96.46} & \cellcolor{gray!30}\textbf{98.29} &\cellcolor{gray!30}- &\cellcolor{gray!30}\textbf{98.27} & \cellcolor{gray!30}59.83 & \cellcolor{gray!30}57.48 & \cellcolor{gray!30}90.25 &\cellcolor{gray!30}83.43 \\

\bottomrule
\end{tabular}
\end{table*}

\subsection{Ablation Study}
To justify the choice of PCA based transformation, we evaluate its effectiveness against alternative pooling strategies. Additional common aggregation functions include $\min$, $\max$, and $\mathrm{mean}$. To select the most effective aggregation function, we construct a validation set consisting of clean images from the dataset of interest and adversarial images generated by various attacks on a specific classifier. As reported in Table~\ref{tab:results_Tiny_minmax}, PCA-based aggregation yields the best separation between clean and adversarial samples averaged on all attacks, and is therefore used in our method.

\begin{table*}[tb]
\centering
\caption{\textbf{Ablation on aggregation functions.} Zero-shot adversarial detection performance (AU-ROC) on validation set 
From tiny-ImageNet and DeiT-Small classifier. \textbf{Bold} - best, \underline{underline} - second best.}
\label{tab:results_Tiny_minmax}
\begin{tabular}{lccccccc|c}
\toprule
Method & FGSM & PGD & AutoAttack & BIM & CW & DeepFool & Square & Avg.\\
\midrule
Max Percentile & \textbf{99.25} & \textbf{98.54} & \textbf{98.63} & \textbf{97.99} & 74.70 & 76.22 & \underline{95.35} & \underline{91.53} \\
Min Percentile & 93.07 & 89.84 & 89.73 & 85.66 & 69.01 & 69.75 & 88.05 & 83.58 \\
Mean Percentile & 98.63 & 97.44 & 97.61 & 96.81 & \underline{75.19} & \underline{76.86} & 94.77 & 91.04\\
PCA Aggregation & \underline{99.24} & \underline{98.16} & \underline{98.39} & \underline{97.76} & \textbf{77.31} & \textbf{78.62} & \textbf{97.84} & \textbf{92.47} \\
\bottomrule
\end{tabular}
\end{table*}

\subsection{Discussion}
The results are significantly better for FGSM, PGD, AutoAttack, and BIM, while the AUC decreases for CW and DeepFool, and we aim to explain this phenomenon. As we have shown, CLIP is sensitive to added noise to the image, so we check how similar the perturbation of each attack is to added white Gaussian noise. Fig.~\ref{fig:all_attacks} qualitatively shows that the perturbations of CW and DeepFool are similar to the edges of the images, while the perturbations of the other attacks resemble noise. For a quantitative measurement, we calculate the correlation between the edges in the perturbations and the edges of the images. The edges were calculated using the Sobel algorithm. The results are shown in Fig~\ref{fig:corr_auc}. The x-axis represents the correlation value, and the y-axis represents the AUC of our method on one dataset and one classifier (Tiny-ImageNet; DeiT-Small). For each attack, there is a circle—the center represents the mean of 200 correlations, and the radius represents the standard deviation. CW and DeepFool perturbations have a higher correlation with the image edges, as expected.

\subsection{Limitations}
Despite its strengths, our method has several limitations. First, its performance depends on the choice of textual prompts and the representation quality of CLIP. In our method, the prompts are chosen manually based on their performance on the validation set. Not all semantically plausible noise-related prompts are suitable for this task, as shown in Figure~\ref{fig:prompt_examples}, where the straightforward general sentence ``A noisy image'' is highly non-discriminative for all attacks. In order to define a dictionary with powerful discriminative capabilities, we aim to select sentences that are both discriminative and low-correlated, as illustrated in Fig.~\ref{fig:clip_prompts}. However, a more systematic procedure for prompt selection could be developed.
There exists a natural trade-off. On the one hand, learning the prompts may lead to overfitting, which contradicts the goal of maintaining a general zero-shot setting. On the other hand, manually selected prompts may not generalize equally well across different attack types. As a compromise, we select semantically relevant prompts with low correlation among them to encourage diversity.
Finally, our approach assumes access to clean training data that is representative of the test distribution. Improving robustness under distribution shift and developing principled prompt selection strategies remain promising directions for future work.

\section{Conclusions}
We proposed a zero-shot adversarial attack detection framework based on prompt-image similarity representations derived from a vision-language model. The method learns a compact representation of clean data using only clean samples and enables test-time detection without prior knowledge of the attack type or the underlying classifier.

By leveraging the expressive representations of vision--language models, the proposed approach can identify subtle perturbations that induce shifts in the embedding space while remaining visually inconspicuous. At the same time, detection requires only a single forward pass per image followed by lightweight similarity-based computations, resulting in an efficient and practical solution. Overall, this work highlights the potential of vision-language models as a general and efficient tool for adversarial detection.

\section*{Acknowledgments}
We would like to acknowledge  
support by the Israel Science Foundation (Grant 1472/23) and by the Ministry of Innovation, Science and Technology (Grant 8801/25).

\FloatBarrier
\bibliographystyle{splncs04}
\bibliography{main}
\end{document}